\documentclass{article}
\usepackage{spconf,amsmath,graphicx}

\usepackage{hyperref}
\usepackage{url}
\usepackage{booktabs}
\usepackage{array}
\usepackage{diagbox}
\usepackage{caption}
\usepackage{subcaption}
\usepackage{bm}
\usepackage{fancyhdr}
\fancypagestyle{firstpage}
{
    \fancyhead[L]{\textsuperscript{\textcopyright} 2021 IEEE. Personal use of this material is permitted. Permission from IEEE must be obtained for all other uses, in any current or future media, including reprinting/republishing this material for advertising or promotional purposes, creating new collective works, for resale or redistribution to servers or lists, or reuse of any copyrighted component of this work in other works.}    
    \fancyhead[R]{}
}

\newcolumntype{C}[1]{>{\centering\arraybackslash}p{#1}}
\title{\vspace{1cm}USING THE OVERLAPPING SCORE TO IMPROVE CORRUPTION BENCHMARKS}
\name{Alfred Laugros$^{\star \dagger}$ \qquad Alice Caplier$^{\star}$ \qquad Matthieu Ospici$^{\dagger}$}
\address{$^{\star}$Universite Grenoble Alpes,\qquad
      $^{\dagger}$Atos}
\begin{document}
%\ninept
%
\maketitle
\begin{abstract}
\thispagestyle{firstpage}
Neural Networks are sensitive to various corruptions that usually occur in real-world applications such as blurs, noises, low-lighting conditions, etc. To estimate the robustness of neural networks to these common corruptions, we generally use a group of modeled corruptions gathered into a benchmark. Unfortunately, no objective criterion exists to determine whether a benchmark is representative of a large diversity of independent corruptions. In this paper, we propose a metric called corruption overlapping score, which can be used to reveal flaws in corruption benchmarks. Two corruptions overlap when the robustnesses of neural networks to these corruptions are correlated. We argue that taking into account overlappings between corruptions can help to improve existing benchmarks or build better ones.
\end{abstract}
\begin{keywords}
Robustness, Benchmark, Corruptions
\end{keywords}
\section{Introduction}
\label{sec:intro}
Neural Networks perform poorly when they deal with images that are drawn from a different distribution than their training samples. Indeed, neural networks are sensitive to adversarial examples \cite{ref_adv}, background changes \cite{background_adv}, and common corruptions \cite{imagenet_c}. Common corruptions are perturbations that change the appearance of images without changing their semantic content. For instance, neural networks are sensitive to noises \cite{add_noise_study}, blurs \cite{blur} or lighting condition variations \cite{cure_tsr}. Contrary to adversarial examples \cite{ref_adv}, common corruptions are not artificial perturbations especially crafted to fool neural networks. They naturally appear in industrial applications without any human interfering, and can significantly reduce the performances of neural networks. 

A neural network is robust to a corruption $c$, when its performances on samples corrupted with $c$ are close to its performances on clean samples. Some methods have been recently proposed to make neural networks more robust to common corruptions \cite{stylized_imagenet,augmix,game_noise}. To determine whether these approaches are effective, it is required to have a method to measure the neural network robustness to common corruptions. The most commonly used method consists in evaluating the performances of neural networks on images distorted by various kinds of common corruptions: \cite{imagenet_c,face_rec_noise,stylized_imagenet,cure_tsr}. In this study, the group of perturbations used to make the robustness estimation is called a corruption benchmark. 

The corruptions used to constitute benchmarks are usually arbitrarily selected \cite{imagenet_c,face_rec_noise,stylized_imagenet,cure_tsr}. No objective criterion is used to determine which corruptions should be included in a benchmark. Yet, we argue that the quality of a robustness estimation varies a lot depending on which corruptions constitute the used benchmark. In this paper, we provide a metric called \textbf{corruption overlapping score}, which can be used to reveal important flaws in benchmarks and help to fix them. Two corruptions overlap when the robustnesses of neural networks towards these corruptions are correlated. The corruption overlapping score estimates to what extent two corruptions overlap. We demonstrate that this metric can be used to determine (1) if a benchmark is unbalanced, i.e. if a benchmark gives too much importance to the robustness to some corruptions compared to others, (2) if a particular corruption is covered by a benchmark, i.e. if being robust to a benchmark implies being robust to a specific corruption. We think that considering the overlappings between corruptions can help to build benchmarks that make better estimations of the neural network robustness.

\section{Related Works}
\textbf{Corruption Benchmarks} Diverse selections of common corruptions have been proposed to build a synthetic corruption benchmark \cite{face_rec_noise,laugros19,stylized_imagenet}. Such benchmarks are used to estimate the robustness of neural networks to natural distribution shifts in the context of object recognition \cite{imagenet_c}, object detection \cite{cc_object_detection}, scene classification \cite{scene_classif} and eye-tracking \cite{cc_gaze}.

\textbf{Corruption Overlappings in Benchmarks.} It has been noticed that fine-tuning a model with camera shake blur helps it to deal with defocus blur and conversely \cite{blur}. The robustnesses to diverse kinds of noises have also been shown to be closely related \cite{laugros19}. In general, it has been shown that the robustnesses to the corruptions that distort the high-frequency content of images are correlated \cite{fourier}. Discussions about the way robustness to synthetic corruptions transfer to robustness to natural distribution shifts are conducted in \cite{many_face_rob,taori_nat} In the context of adversarial examples, it is known that the robustness towards one adversarial attack can be correlated with the robustness to another attack \cite{multi_adv_rob}. So, it is generally recommended to evaluate the adversarial robustness with attacks that are not too correlated in terms of robustness \cite{evaluating}. The experiments carried out in this paper suggest that this recommendation should also be followed in the context of common corruption robustness estimation.

\section{The Corruption Overlapping Score \label{sec:overlapping_score}}
\thispagestyle{empty}
We consider that two corruptions overlap when the robustness to one of these corruptions is correlated with the robustness to the other corruption. In this section, we propose a methodology to estimate to what extent two corruptions overlap.

\textbf{The Robustness Score.} To determine whether two corruptions overlap, we first need to introduce a metric called the robustness score. This score gives an estimation of the robustness of a model $m$ to a corruption $c$. It is computed with the following formula: $R^m_c = \frac{A_c}{A_{clean}}$.

$A_{clean}$ is the accuracy of $m$ on an uncorrupted test set and $A_c$ is the accuracy of $m$ on the same test set corrupted with $c$. The higher $R^m_c$ is, the more robust $m$ is.

\textbf{The Corruption Overlapping Score.} We consider two neural networks $m1$ and $m2$ and two corruptions $c1$ and $c2$. $m1$ and $m2$ are identical, and trained with exactly the same settings except that their training sets are respectively augmented with the corruptions $c1$ and $c2$. A standard model is trained the same way but only with non-corrupted samples. We propose a method to measure to what extent $c1$ and $c2$ overlap. The idea of the method is to see if a data augmentation with $c1$ makes a model more robust to $c2$ and conversely. To determine this, $m1$, $m2$, and a test set are used to compute the following expression:

\begin{equation}
(R^{m2}_{c1} - R^{standard}_{c1}) + (R^{m1}_{c2} - R^{standard}_{c2})  \label{overlapping_terms}
\end{equation}

The first term of (\ref{overlapping_terms}) measures if a model that fits exactly $c2$ is more robust to $c1$ than the standard model. Symmetrically, the second term measures if a model that fits exactly $c1$ is more robust than the standard model to $c2$. The more fitting exactly $c1$ implies being robust to $c2$ and reciprocally, and the more we can suppose that the robustnesses to $c1$ and $c2$ are correlated in practice. In other words, the expression (\ref{overlapping_terms}) gives an estimation of the overlapping between $c1$ and $c2$. To be more convenient, we would like to build a corruption overlapping score equal to 1 when $c1=c2$, and equal to 0 when the robustnesses to $c1$ and $c2$ are not correlated at all. We propose a new expression that respects both conditions:

\begin{small}
\begin{equation}
O_{c_1,c_2} = \max\{0,\frac{1}{2}*\left(\frac{R^{m1}_{c2}-R^{standard}_{c2}}{R^{m2}_{c2}-R^{standard}_{c2}}   +     \frac{R^{m2}_{c1}-R^{standard}_{c1}}{R^{m1}_{c1}-R^{standard}_{c1}}\right)\} \label{overlaping_expr}
\end{equation}
\end{small}

The expression (\ref{overlaping_expr}) is a normalized version of (\ref{overlapping_terms}). It measures the overlapping between two corruptions while respecting the conditions mentioned above. Indeed, if a data augmentation with $c1$ does not increase the robustness to $c2$ at all and conversely, then the ratios in (\ref{overlaping_expr}) are null or negative, so the whole overlapping score is maximized to zero. Besides, when $c1=c2$,  $R^{m1}_{c2} = R^{m2}_{c2}$ and  $R^{m2}_{c1} = R^{m1}_{c1}$, so both ratios of (\ref{overlaping_expr}) are equal to 1. Then, $O_{c_1,c_2}=1$ when $c1$ and $c2$ completely overlap.

\begin{figure*}
\centering
\includegraphics[scale=0.25]{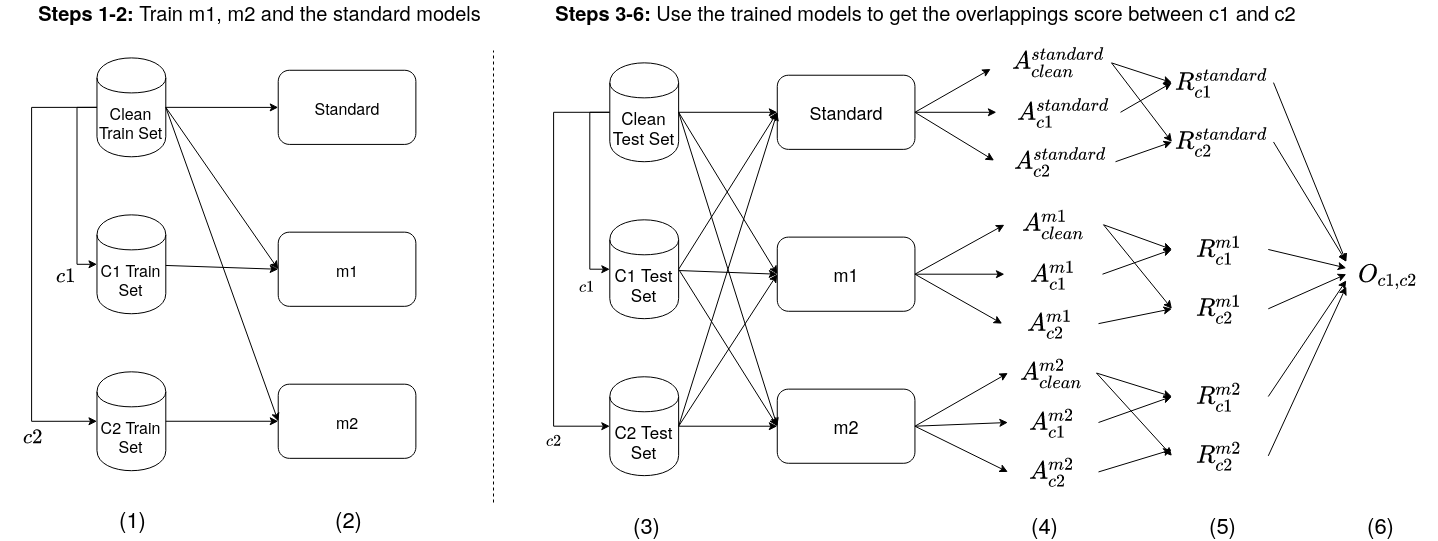}
\caption{Methodology used to compute the overlapping score between two corruptions $c_1$ and $c_2$. \label{fig:overlaping_diagram}}
\end{figure*}

\textbf{How to compute an overlapping score.} To get the overlapping score between $c1$ and $c2$, we follow the method illustrated in Figure \ref{fig:overlaping_diagram}. This method has six steps, and requires to have a training set, a test set and three untrained models that share the same architecture ($m1$, $m2$ and $standard$). Step (1), consists in using the corruptions $c1$ and $c2$ to get two training sets, each corrupted with one corruption. Then, the obtained corrupted sets are used to train the models $m1$ and $m2$ in step (2). The standard model is also trained during this step but only with non-corrupted samples. In step (3), similarly to step (1), we use $c1$ and $c2$ to get two corrupted versions of the test set. The accuracies of the three models on the three test sets are computed in step (4). The scores obtained are used in step (5), to get the robustness scores of each model for the corruptions $c1$ and $c2$. The results obtained are used to compute the overlapping score between $c1$ and $c2$ in step (6). To be meaningful, an overlapping score should be computed with models that converged adequately on their training sets. In all experiments, we verify that all trained models converge.

\section{Experiments \label{sec:imagenoc_construction}}
\thispagestyle{empty}
\textbf{Experimental Set-up.} For every training, we use the following parameters. The used optimizer is SGD with a momentum of 0.9. The used cost function is a cross-entropy function, with a weight decay set to $10^{-4}$.  Models are trained for 40 epochs with a batch size of 256. The initial learning rate is set to 0.1 and is divided by 10 at epoch 20 and 30. We use the CE score metric \cite{imagenet_c} to measure the robustness of models to corruptions in all our experiments. The CE score is an error rate computed with corrupted samples, weighted with the AlexNet error rate. It is useful to compare the robustness of models that have a different error rate on clean samples.

\subsection{Overlappings in ImageNet-C}
We propose to use the corruption overlapping score to study ImageNet-C: a benchmark commonly used to measure the robustness of ImageNet classifiers to common corruptions \cite{imagenet_c}. It is built on fifteen common corruptions called \textit{Gaussian noise, shot noise, impulse noise, defocus blur, glass blur, motion blur, zoom blur, snow, frost, fog, brightness, contrast, elastic, pixelate}, and \textit{jpeg compression}. We use the methodology presented in Figure \ref{fig:overlaping_diagram} to compute the overlapping scores between all the ImageNet-C corruptions. The dataset used is ImageNet-100: a subset of ImageNet that contains every tenth ImageNet class by WordNetID order \cite{wordnet_id}. All images are resized to the 224x224 format, and randomly horizontally flipped with a probability of 0.5 during trainings. When a model is trained with data augmentation, half of the images of each training batch are transformed with a corruption, while the other half is not corrupted. In Figure \ref{fig:overlapping_scores}, we display the overlapping scores computed with the ResNet-18 and DenseNet-121 architectures. We observe that both matrices are very similar, which suggests that our overlapping metric is network architecture independent. Computing all these overlapping scores required to train thirty models and took two weeks with a single GPU Nvidia Tesla V100.

The overlapping score predicts that several groups of ImageNet-C corruptions such as noises are very correlated in terms of robustness. In other words, the metric predicts that being robust to one of the ImageNet-C noises implies being robust to the other ones. Interestingly, we observe that the corruptions that damage the textures in images (blurs, noises, pixelate and jpeg) significantly overlap. This is consistent with the experiments carried out in \cite{fourier}, where it is argued that the robustnesses of neural networks to corruptions that alter high-frequency information of images are correlated.

\subsection{Corruption Overlapping and Benchmark Balance}
We consider two corruptions $c1$ and $c2$ in a benchmark $bench$. We say that $bench$ is unbalanced, when the robustness to $bench$ (the mean robustness computed with the corruptions in $bench$) is more correlated with the robustness to $c1$ than with the robustness to $c2$ or vice versa. For instance, we show in this section that ImageNet-C is unbalanced because the robustness to this benchmark is more correlated with the robustness to blurs than with the robustness to brightness variations. Yet, we think that being robust to different kinds of blurs is not more valuable than being robust to lighting condition variations. So, being unbalanced is in general not a desirable property because it makes benchmarks give biased estimations of neural network robustness.

\begin{figure*}
\centering
\includegraphics[width=\textwidth]{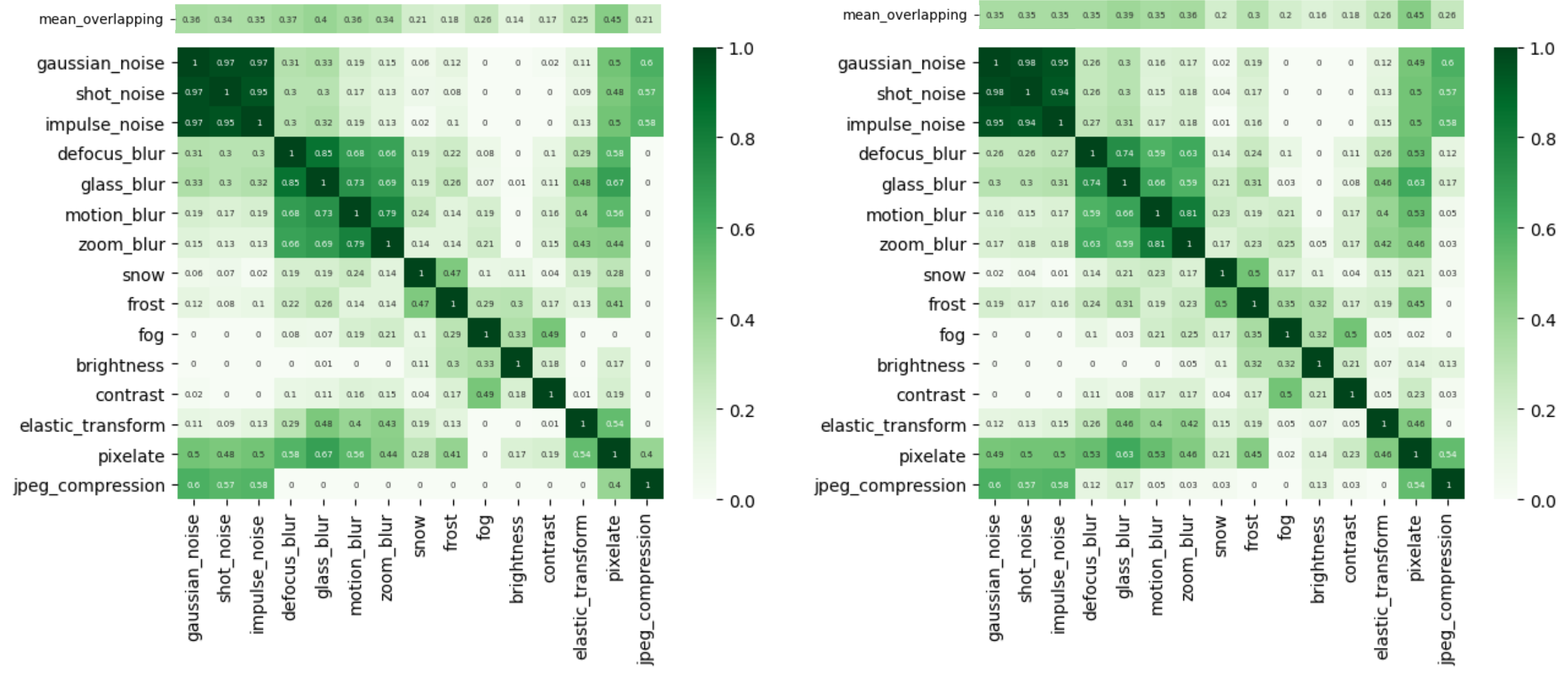}
\caption{The overlapping scores of ImageNet-C computed with the ResNet-18 (left) and DenseNet-121 (right) architectures \label{fig:overlapping_scores}}
\end{figure*}

We propose a simple method based on the overlapping score to determine if a benchmark is unbalanced. If $c1$ overlaps more with the other corruptions of $bench$ than $c2$, then being robust to the corruptions of $bench$ is more correlated with being robust to $c1$ than being robust to $c2$. So, if we show that a corruption contributes more to the total overlapping of a benchmark than another corruption, then this benchmark is unbalanced. Following this idea, we compute the mean overlapping associated with each ImageNet-C corruption. The results are displayed in Figure \ref{fig:overlapping_scores}. It appears that some corruptions contribute more to the total overlapping than others, then ImageNet-C is probably unbalanced. 

To verify this, we proceed to the following experiment: we consider a first group $set_{1}$ of eight corruptions, constituted with the three ImageNet-C noises, its four blurs and its \textit{pixelate} corruption. This group corresponds to the corruptions that contribute the most to the overlappings in ImageNet-C. $set_{2}$ corresponds to the group containing the remaining ImageNet-C corruptions. We compute the mean CE scores of the standard torchivsion pretrained ResNet-50 for both $set_{1}$ and $set_{2}$. In Table \ref{tab:sota_on_nocs}, we compare these scores with the ones obtained with four models that have been shown to be robust to ImageNet-C called \textit{SIN+IN} \cite{stylized_imagenet}, \textit{ANT\textsuperscript{3x3}} \cite{game_noise}, \textit{Augmix} \cite{augmix} and \textit{DeepAugment} \cite{many_face_rob}. Indeed, these four models are more robust than the standard ResNet-50, but we observe that the robustness gain is much higher for $set_{1}$ than $set_{2}$. \textit{ANT\textsuperscript{3x3}}, \textit{Augmix} and \textit{DeepAugment} have been designed to resist common corruptions, and ImageNet-C have been used as a reference by these models to estimate their robustness to common corruptions. The overlapping scores predict that ImageNet-C gives more importance to the robustness to corruptions such as blurs than to the robustness to other corruptions such as brightness variations. So, it is not surprising to see that the studied models are more robust to blurs than brightness variations: this is a direct consequence of the unbalance of ImageNet-C.

\begin{table}
\begin{center}
\begin{small}
\caption{The CE scores of various models computed with the \textit{border}, \textit{obstruction}, $set_{1}$ and $set_{2}$ corruptions. The scores in brackets indicate the CE gains compared to the standard ResNet50. \label{tab:sota_on_nocs}}
\begin{tabular}{p{12mm}||C{17mm}C{18mm}|C{8mm}C{7mm}}
\toprule
{} & mean\_CE\_Set1 &  mean\_CE\_Set2 &  Border &  Obstr  \\
\midrule
Standard & 81 (0) & 73 (0)  &   53 &   63 \\
SIN+IN & 71 (-10) & 68 (-5)  &   56 &   69 \\
Augmix  & 66 (-15) & 65 (-8) & 50 &   63  \\
ANT\textsuperscript{3x3}  & 60 (-21) &  68 (-5) &   58 &  71 \\
DeepAug  & 59 (-22) & 63 (-10) &  60 &   72  \\
\bottomrule
\end{tabular}
\end{small}
\end{center}
\end{table}

\subsection{Corruption Overlapping and Covered Corruptions} 
\thispagestyle{empty}
We consider that a corruption $c$ is covered by a benchmark when being robust to this benchmark implies being robust to $c$. In practice, we want benchmarks to cover as many corruptions as possible. For instance, if a contrast loss corruption is not covered by a benchmark, then a model could be robust to this benchmark without being robust to contrast loss. So this benchmark omits an important kind of corruption. Then, finding corruptions that are not covered by a benchmark reveals weaknesses of this benchmark and suggests improvements. We propose a simple method based on the overlapping score to show that a corruption $c$ is not covered by a benchmark $bench$. The idea is to compute all the overlapping scores between $c$ and the $bench$ corruptions. If these scores are all equal to zero, then being robust to $bench$ should not imply being robust to $c$; namely $c$ is not covered by $bench$.

To confirm the validity of our approach, we model a group of common corruptions such as image shear, rain modeling or pixel quantization. We compute the overlapping scores between all these corruptions and all the ImageNet-C corruptions. We found two corruptions that do not overlap at all with any of the ImageNet-C corruptions that we call \textit{border} and \textit{obstruction}. The \textit{border} corruption replaces the values of the pixels at the edge of images by a unique value from 0 to 1 drawn from a uniform distribution. For each corrupted image, the thickness of the masked area is randomly chosen between 10 and 45 pixels. \textit{Obstruction} selects a randomly positioned square area in images, and replaces the value of the pixels in this area by a unique value drawn from a uniform distribution from 0 to 1. The size of the edge of the square is randomly chosen from 50 to 120 pixels. We compute the CE scores of the four models robust to ImageNet-C with these two corruptions, and we compare their results with the ones obtained using the standard ResNet-50 in Table \ref{tab:sota_on_nocs}. We see that \textit{SIN+IN}, \textit{ANT\textsuperscript{3x3}} and \textit{DeepAugment} are less robust than the standard ResNet-50 to the two corruptions. Then, in practice, being robust to ImageNet-C does not imply being robust to \textit{border} and \textit{obstruction}, i.e. these two corruptions are not covered by ImageNet-C.

\subsection{Conclusion}
Selecting the synthetic corruptions to constitute a benchmark is not an easy task. We demonstrated that corruption benchmarks can be unbalanced, and they may not provide any robustness guarantee on some kinds of common corruptions. Consequently, objective criteria that help to find relevant sets of corruptions to constitute benchmarks are required. We argue that such criteria could be based on the overlapping score. For instance, a good practice when building a benchmark could be to \textit{add a new corruption $c$ to a benchmark, only if $c$ does not overlap with any corruptions of this benchmark.} By using this criterion, we make sure that the corruptions in a benchmark are not correlated in terms of robustness, which implies to build a balanced benchmark. Besides, this criterion also guarantees that the benchmark does not already provide a robustness guarantee on the corruptions that are added to it. So, the more corruptions that fit this criterion are found and included into the benchmark, the more this benchmark covers a wide range of corruptions. More refined criteria could be proposed and we intend to investigate this in further works.

\pagestyle{empty}
\bibliographystyle{IEEEbib}
\bibliography{strings,refs}
\end{document}